\documentclass[journal]{IEEEtran}

\usepackage{bbding}
\usepackage{pifont}
\usepackage{threeparttable}
\usepackage{amsfonts}
\usepackage{multirow}
\usepackage{tabularx}
\usepackage{booktabs}

\ifCLASSINFOpdf
  \usepackage[pdftex]{graphicx}
\else
  \usepackage[dvips]{graphicx}
\fi
\usepackage{amsmath}
\usepackage{algorithm, algorithmicx}
\usepackage{algpseudocode}

\begin{document}
%
\title{Collaborative Knowledge Infusion \\ for Low-resource Stance Detection}
%
\author{Ming Yan,
        Joey Tianyi Zhou~\IEEEmembership{Member,~IEEE,}
        and Ivor W. Tsang~\IEEEmembership {Fellow,~IEEE.} \\ 
\thanks{Ming Yan, Joey Tianyi Zhou, and Ivor W. Tsang are with Centre for Frontier AI Research (CFAR), and Institute of High Performance Computing (IHPC), Agency for Science Technology and Research (A*STAR), 1 Fusionopolis Way, Singapore, 138632, Singapore e-mail: {Yan\_Ming,Joey\_Zhou, Ivor\_Tsang}@cfar.a-star.edu.sg}} 

\maketitle

\begin{abstract}
Stance detection is the view towards a specific target by a given context (\textit{e.g.} tweets, commercial reviews). Target-related knowledge is often needed to assist stance detection models in understanding the target well and making detection correctly. However, prevailing works for knowledge-infused stance detection predominantly incorporate target knowledge from a singular source that lacks knowledge verification in limited domain knowledge. The low-resource training data further increases the challenge for the data-driven large models in this task. To address those challenges, we propose a collaborative knowledge infusion approach for low-resource stance detection tasks, employing a combination of aligned knowledge enhancement and efficient parameter learning techniques. Specifically, our stance detection approach leverages target background knowledge collaboratively from different knowledge sources with the help of knowledge alignment. Additionally, we also introduce the parameter-efficient collaborative adaptor with a staged optimization algorithm, which collaboratively addresses the challenges associated with low-resource stance detection tasks from both network structure and learning perspectives. To assess the effectiveness of our method, we conduct extensive experiments on three public stance detection datasets, including low-resource and cross-target settings. The results demonstrate significant performance improvements compared to the existing stance detection approaches.
\end{abstract}


\IEEEpeerreviewmaketitle

\section{Introduction}
\IEEEPARstart{S}{tance} detection is the view towards a specific target with a given context, such as tweets or commercial reviews. Typically, those given contexts in stance detection tasks are mostly short-length contexts, which makes it challenging to predict the target's stance for the data-driven detection models with such limited information.
Large pretrained language models (PLMs) are becoming the default backbone to enhance the stance detection model with learned commonsense knowledge, leading to great success in this field~\cite{Ghosh_BERT_Stance, Sen_BERT_Stance}. To further enrich the knowledge of targets, the straightforward approach is to incorporate the target-related background knowledge as extra supplementary knowledge for the pretrained stance detection model, which has been shown to substantially improve model performance~\cite{kawintiranon-singh-2021-knowledge, he-etal-2022-infusing}. In detail, those works infuse explicitly knowledge individually through knowledge graph~\cite{liu-etal-2021-enhancing, agarwal-etal-2021-knowledge}, Wikipedia~\cite{xu2022human, KEPLER2019}, generative knowledge~\cite{lin-etal-2021-bertgcn, wei2022chain}, leveraging PLMs' knowledge feature learning and representation capability by fine-tuning entire models' parameters. 
However, those knowledge-infuse solutions are quite inefficient in fine-tuning large PLM backbones on the limited training data. For instance, the few/zero-shot stance detection dataset VAST~\cite{allaway-mckeown-2020-zero} has very limited training data or even no training data for each target. Besides the low-resource challenge, unbalanced dataset distribution is another challenge for the stance detection task, leading the training trajectory to fall into the local minima. Last but not least, we find some background knowledge is not always infused correctly in the knowledge infusion process. This is because a single knowledge source in previous works can not fully cover and support enough knowledge for diverse targets~\cite{he-etal-2022-infusing}. For example, the target ``breaking the law'' in Wikipedia is erroneously linked to a heavy metal music song rather than its ground truth definition of engaging in activities contrary to the law.

To address the aforementioned challenges, we propose a novel collaborative knowledge-infused stance detection method for training the large detection model in the low-resource setting efficiently. Specifically, we introduce a retrieval-based knowledge verifier that mitigates incorrect knowledge infusion by selecting the high-semantic background knowledge from different knowledge sources, rather than relying on a single knowledge source. 
Furthermore, we present a trainable collaborative adaptor integrated into PLMs to enable efficient parameter learning in low-resource stance detection tasks. 
Concretely, the collaborative adaptor freezes the parameter weights of large PLM and fine-tunes the parameter-efficient adaptor only, which alleviates the over-fitting effects on large PLM in low-resource scenarios. However, we empirically find that intuitively adding adaptors into PLM may lead to unstable training in the new stance detection tasks. We think the initialized weights of the collaborative adaptor can not work well with the pretrained PLMs in the early fine-tuning stage of new tasks. Moreover, the unbalanced data distribution further impacts the stable training. So we design a staged optimization algorithm for the adaptive model training in unbalanced distributions. The primary objective of the first optimization stage is to prevent the training trajectory from converging to a local minimum leading to unexpected performance. In the second stage, our model introduces a weighted cross-entropy loss to balance the biased stance categories and further improve the model performance in low-resource stance detection tasks. In other words, we progressively used label smooth (stage-1) and weighted loss (stage-2) separately to reduce the over-fitting effects in our low-resource stance detection tasks, which is different from traditional optimization paradigms using those two in the whole training process.

We conducted extensive experiments on three public stance detection datasets, encompassing the low-resource stance detection, and cross-target stance detection tasks. Experimental results demonstrate the superior performance of our method compared to state-of-the-art approaches across all stance detection tasks. The contributions of our work are summarised as follows, 
\begin{itemize}
    \item We introduce collaborative knowledge verification to assist the detection model in selecting more semantic-related knowledge from different knowledge sources. To our knowledge, this is the first work to infuse verified knowledge into the knowledge enhancement stance detection task.
    \item We introduce a collaborative adaptor to selective learning features in an efficient way for the low-resource setting. It contains three sub-components which are architecturally located in different positions of the backbone model, learning different features collaboratively.  
    \item To alleviate the unbalanced effects of low-resource stance detection tasks, we also provided a staged optimization algorithm to improve the training efficiency in large PLMs. Experiments showed the superiority of our method in different low-resource settings and outperformed state-of-the-art approaches on three public stance detection datasets.
\end{itemize}

\section{Related Works}
\subsection{Knowledge Enhancement}
Knowledge enhancement increases the capabilities in thinking, understanding, and reasoning for the data-driven models beyond the original training data. In recent years, there has been a growing trend in infusing external-specific knowledge as complementary knowledge to the large pretrained models~\cite{zhu2022knowledge}. Depending on the infused knowledge, knowledge infusion methods can be broadly categorized into structured-knowledge infusion (\textit{e.g.}, knowledge graph) and unstructured-knowledge infusion (\textit{e.g.}, Wikipedia, text corpus).

Domain-specific experts typically collect structured knowledge and encompass well-organized and rich knowledge. For instance, CKE-Net~\cite{liu-etal-2021-enhancing} utilizes the structured knowledge base (ConceptNet) to enhance its model's common sense knowledge in zero/few-shot stance detection tasks. Similarly, K-BERT~\cite{Liu_Zhou_Zhao_Wang_Ju_Deng_Wang_2020} incorporates domain knowledge through entity triplets obtained from the knowledge graph. Other methods like JAKET~\cite{Yu_Zhu_Yang_Zeng_2022}, ERNIE~\cite{zhang-etal-2019-ernie} and Entity-as Experts~\cite{fevry-etal-2020-entities} also infuse knowledge from knowledge bases through grounding knowledge with entity linking technologies. Structured knowledge provides well-organized and domain-specific knowledge for specific targets in stance detection tasks. However, its utility is limited by the pre-defined scope of available knowledge, which may not cover all targets encountered in practical scenarios.

In contrast, unstructured knowledge offers more flexibility and can be easily collected from a wide range of diverse domains. For instance, the VAST~\cite{allaway-mckeown-2020-zero} dataset introduces thousands of diverse targets that mostly can not be found in the well-constructed structured knowledge. To incorporate unstructured knowledge into the stance detection models, WS-BERT~\cite{he-etal-2022-infusing} directly infuses external knowledge from Wikipedia as its inputs to pre-trained models for stance detection in the VAST dataset. Another knowledge infusion paradigm is that finetune PLMs on the specific domain corpus to embed the domain-specific knowledge, as demonstrated by Sci-BERT~\cite{beltagy-etal-2019-scibert}, Bio-BERT~\cite{kanakarajan-etal-2019-biobert}, BERTweet~\cite{nguyen-etal-2020-bertweet}. In addition to domain-specific finetuning, Self-talk~\cite{shwartz-etal-2020-unsupervised} offers another interesting solution by exploring knowledge from its own training corpus with hand-crafted prompts, enhancing language model learning with task-related knowledge. Furthermore, DDP~\cite{karpukhin-etal-2020-dense} and K-Former~\cite{10.1007/K_Former} present the retrieval-based knowledge infusion methods by retrieving knowledge from feature pools and online websites, respectively. Nevertheless, more efforts are still needed to collaborate structured and unstructured knowledge together in a correct and efficient manner for large PLM-based models, particularly in low-resource tasks.

\subsection{Stance Detection}
Stance detection refers to the identification of attitudes toward a specific context or topic, typically framed as a stance classification problem (pros,  cons, neutral) for the neural network-based models. Stance detection encompasses various tasks depending on the specific topics involved, including rummer stance detection~\cite{10.1145/3369026}, fake news stance detection~\cite{ALDAYEL2021102597}, disinformation/misinformation stance detection~\cite{hardalov-etal-2022-survey}, multi/cross-language stance detection~\cite{mohtarami-etal-2019-contrastive, zotova-etal-2020-multilingual} and zero-shot stance detection~\cite{allaway-mckeown-2020-zero}, \textit{etc}. In this study, we focus on in stance detection tasks of background knowledge infusion and low-resource training. 
To infuse background knowledge, existing approaches try to incorporate knowledge from different sources. For instance, CKE-Net~\cite{liu-etal-2021-enhancing} introduces target-related knowledge from ConceptNet, which is trained on the common sense knowledge graph. Similarly, BS-GGCN~\cite{luo-etal-2022-exploiting} simplifies the whole Concept-Net graph to a compact sentence-related graph, enabling more efficient knowledge embedding for stance detection. Moreover, WS-BERT~\cite{he-etal-2022-infusing} leverages the background knowledge from Wikipedia pages as its additional input to improve the model performance in stance detection. 
Regarding the low-resource challenge, STCC~\cite{liu-etal-2022-target} employs contrastive learning to enhance target representation in low-resource stance detection tasks. Different from STCC building contrastive examples in the existence of the target, Joint-CL~\cite{liang-etal-2022-jointcl} builds contrastive examples from the prototype graph representation of the target's link, which further improves the model performance on the unseen targets. While most of those works solve the knowledge infusion and low-resource separately, it is essential to consider two challenges together to improve the performance of stance detection models.

\begin{figure*}[!t]
\centering
\includegraphics[width=0.95\linewidth]{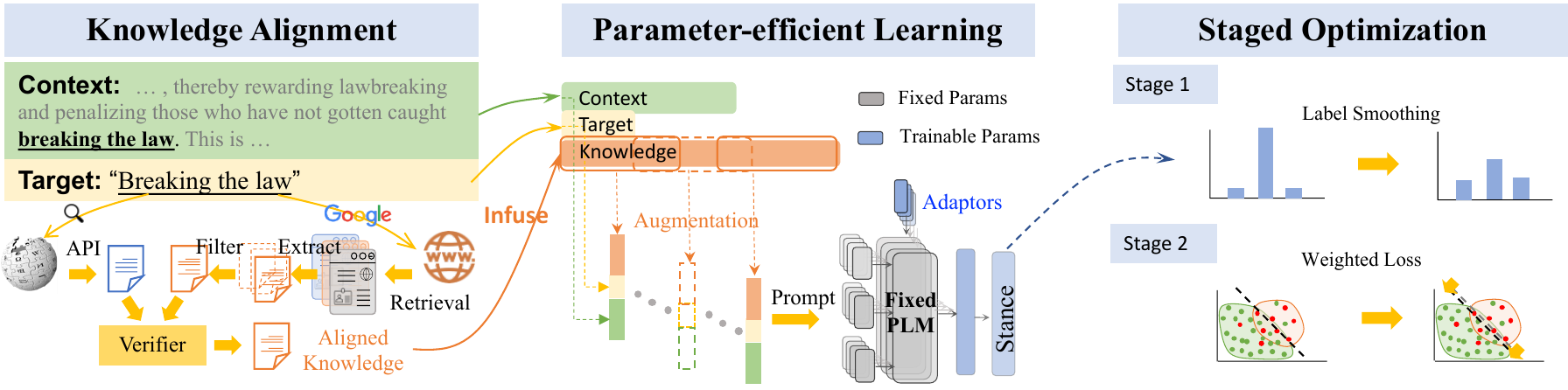}
\caption{Overview of our stance detection architecture: knowledge alignment, parameter-efficient learning, and staged optimization. Knowledge alignment collaboratively selects semantic similar knowledge of the target from different knowledge sources. Parameter-efficient learning introduces the collaborative adaptor and knowledge augmentation into the stance detection model to perform low-resource learning. Staged optimization algorithm optimizes the classifier with label smoothing, then pushes the classifier edge aligning to data distribution with the help of weighted loss.}
\label{fig2_overview}
\end{figure*}

\section{Methodology}
Before delving into our methodology, we define the notations of the stance detection task as follows: 
Given a context set $\mathbb{C}$ with elements $c_i$ where $i = 1, 2, ..., n_c$, and a target set $\mathbb{T}$ with element $t_j$ where $j = 1, 2, ..., n_t$. The stance detection task is formulated to predict the stance $y$ that maximizes  $P(y|\mathbb{C}, \mathbb{T}\})$, where the stance set is $Y=\{\mathtt{pros}, \mathtt{cons}, \mathtt{neutral} \}$. Regarding the knowledge enhancement stance detection task, its objective is defined as maximizing $P(y|\{\mathbb{C}, \mathbb{T}, \mathbb{K} \})$, where $\mathbb{K}$ denotes the infused knowledge that assists in the stance detection task.

The overview of our proposed methodology, as illustrated in Figure~\ref{fig2_overview}, contains three modules: 
(1) Knowledge alignment, which aims to collaborative select semantic target knowledge from structured and unstructured knowledge sources. 
(2) Parameter-efficient learning, which involves collaborative adaptor and knowledge augmentation to enhance model performance in low-resource settings. 
(3) Staged optimization algorithm, which further refines the adaptive model through strategies such as label smoothing and weighted loss.

\subsection{Knowledge Alignment}
\begin{figure}[!ht]
\centering
\includegraphics[width=0.95\linewidth]{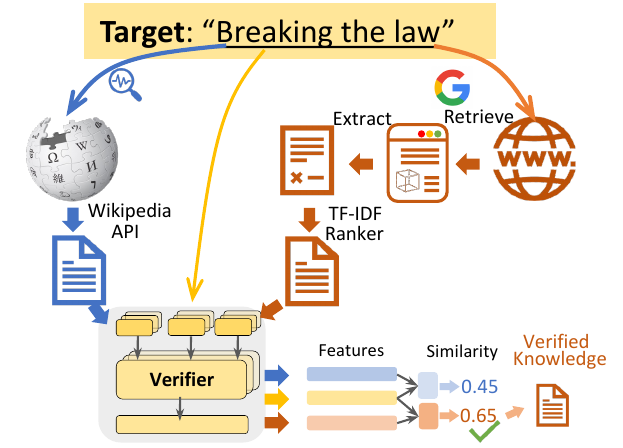}
\caption{Knowledge alignment. The target's collaborative knowledge is the knowledge with a higher semantic similarity score from Wikipedia or the Internet. The target's Wikipedia knowledge is obtained by Wikipedia's API. The target's knowledge from the Internet is obtained by Google retrieval.} 
\label{fig3_knowledgeVerifier}
\end{figure}

We introduce a knowledge alignment module to assist the stance detection model infuse target-related background knowledge correctly, particularly for the targets that lack matching items in singular knowledge source (\textit{e.g.}, Wikipedia). To help the unmatched targets infuse knowledge out of Wikipedia, we incorporate retrieval knowledge from the Internet, specifically through Google search, as an additional knowledge source. Thus, our approach adopts a multi-source knowledge infusion paradigm across structured Wikipedia and unstructured Internet, which selects the semantic similar knowledge as the collaborative knowledge $\mathbb{K}$ to align to the target $\mathbb{T}$ from multiple knowledge sources. Figure~\ref{fig3_knowledgeVerifier} illustrates our knowledge alignment module, as follows,

In the paradigms of knowledge enhancement stance detection, detection models mostly infuse extra knowledge through the target ($\mathbb{T}$) rather than the given context ($\mathbb{C}$). There are two reasons. Firstly, the stance detection task aims to identify the stance ($Y$) of the target, which may not be explicitly mentioned in the given context. Consequently, the target contains more information compared to the given context. Secondly, the context is typically long and complex, making it challenging to locate target-related information for infusing background knowledge ($\mathbb{K}$). In our proposed collaborative knowledge infusion approach (Figure~\ref{fig3_knowledgeVerifier}), we collaboratively incorporate the background knowledge into the detection model by retrieving the target-related knowledge from Wikipedia and the Internet.

For structured Wikipedia knowledge, we utilize the target as the keyword to retrieve background knowledge through Wikipedia's API\footnote{https://github.com/goldsmith/Wikipedia}. This API returns a summary of the matched Wikipedia page. In cases where is no match for a target, we follow the setting of \cite{he-etal-2022-infusing} and consider the target itself as the knowledge without introducing additional information. 

For unstructured Internet knowledge, we retrieve the target-related web pages by using the searching prompt \textit{``What is the meaning/definition of $\mathtt{TARGET}$ $(\mathbb{T})$''}? as the search term for the Google search engine. Subsequently, we select the top three pages from the Google search results and employ BeautifulSoup~\footnote{https://git.launchpad.net/beautifulsoup} to parse the \textit{HTML} contents of these pages into candidate passage lists ($\mathbb{D}$). The next step involves filtering out unrelated contexts from the candidate passage lists, as web pages often contain noise and extraneous information. To accomplish this, we utilize \textit{TF-IDF} ranker to identify and exclude noisy passages from a long list of candidate passages. 

\begin{equation}
     \mathtt{TFIDF}(\mathbb{T},\mathbb{D}) = \mathtt{TF}(\mathbb{T},\mathbb{D}) \cdot \mathtt{IDF}(\mathbb{T}),
\end{equation}

Once the knowledge related to the target has been collected from Wikipedia and the Internet, we introduce the knowledge verifier to select more accurate knowledge from multiple sources as the infused knowledge. Knowledge verification involves feature encoding and feature similarity comparison, that selects semantic similar knowledge among different knowledge sources. Concretely, we employ Sentence-BERT~\cite{reimers-gurevych-2019-sentence} to encode the target $\mathbb{T}$ and its corresponding knowledge (Wikipedia knowledge $\mathbb{K}_w$ and Internet knowledge $\mathbb{K}_g$) into embedding features ($\mathbb{T}^{em}, \mathbb{K}_w^{em}, \mathbb{K}_g^{em}$), as follows, 

\begin{equation}
\label{Eq:embedding}
   \left\{ \mathbb{T}^{em}, \mathbb{K}_g^{em}, \mathbb{K}_w^{em}\right\}  = Stance( \left\{\mathbb{T}, \mathbb{K}_w, \mathbb{K}_g \right\}), \\
\end{equation}

Subsequently, we compute the semantic similarity between the stance target and different knowledge using the classical cosine similarity:

\begin{align}
    S(\mathbb{T}^{em},  \mathbb{K}_w^{em}) & = \frac{\mathbb{T}^{em},  \mathbb{K}_w^{em}}{ ||\mathbb{T}^{em}|| \times ||\mathbb{K}_w^{em}|| }, \\
    S(\mathbb{T}^{em},  \mathbb{K}_g^{em}) & = \frac{\mathbb{T}^{em},  \mathbb{K}_g^{em}}{ ||\mathbb{T}^{em}|| \times ||\mathbb{K}_g^{em}|| }, 
\end{align}

Finally, we select the knowledge with the highest semantic similarity as the collaborative knowledge $\mathbb{K}$ to be infused into our model, which is expressed as follows: 

\begin{equation}
    \mathbb{K} = argmax \{S(\mathbb{T}^{em},  \mathbb{K}_w^{em}), S(\mathbb{T}^{em},  \mathbb{K}_g^{em})\}, 
    \label{Eq:cosineSimilirity}
\end{equation}

By collaborative integration of this verified knowledge, our stance detection model allows for the inclusion of more reliable knowledge in stance detection tasks. This knowledge infusion manner expands the scope of target knowledge by incorporating information from both structured and unstructured knowledge sources. As a result, it addresses the limitations associated with relying on a single knowledge source, including the issues of out-of-scope knowledge and false infusions.

\subsection{Efficient Parameter Learning}
To enable efficient parameter learning for low-resource stance detection tasks, our approach introduces collaborative adaptor and knowledge augmentation into our PLM-based stance detection model. Collaborative adaptor significantly reduces parameters compared to fine-tuning the entire model for low-resource stance detection tasks. Additionally, our collaborative adaptor learns diverse feature representations by leveraging the collaboration of multiple adaptors. To address the input-length limitation of stance models when incorporating collaborative knowledge contexts, we also introduce knowledge augmentation, which helps overcome the constraints imposed by the PLM's input length. 

Suppose our stance detection model is parameterized by the fixed pretrained PLM backbone $W$ and collaborative trainable adaptors $\Delta W$. The backbone model can be a generic language model BERT or RoBERTa with Transformer architecture. In our low-resource stance detection task, the objective of the knowledge-infused stance detection task is formulated as follows:

\begin{equation}
   J = \max_{\Delta W} \sum^{\mathbb{T},\mathbb{C}} \sum^{Y} log(P_{W+\Delta W}(Y|\{ \mathbb{C},\mathbb{T}, \mathbb{K})).
   \label{Eq6:objective}
\end{equation} 

During the training process, our approach keeps the parameters of the pretrained model fixed and trains the collaborative adaptor on the low-resource stance detection dataset. The fixed-weight setting prevents the catastrophic forgetting problem and mitigates the challenges associated with training large models on limited training data ($\mathbb{C}$). Moreover, the collaborative adaptor has significantly fewer parameters than the PLMs $|\Delta W| << |W| $ (\textit{e.g.}, the prefix-tuning adaptor of our collaborative adaptor only has approximately $0.01\%$ of the parameters $W$ in BERT). This reduction in parameters greatly alleviates the data dependency for large model training. 

\begin{figure}[!ht]
\centering
\includegraphics[width=0.8\linewidth]{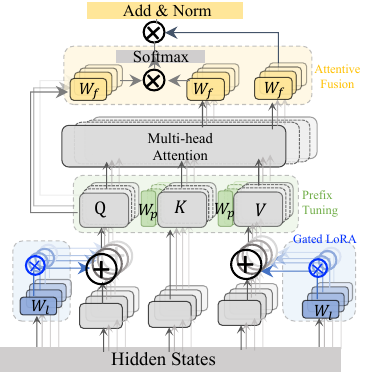}
\caption{Overview of collaborative adaptor in efficient-parameter learning. $Q, K, V$ are the query, key, and value of the Transformer module respectively. LoRA denotes the low-rank adaptor, and the gate is a controller for LoRA.}
\label{fig4_Adaptor}
\end{figure}

The motivation of our collaborative adaptor is to introduce multiple adaptive modules that collaboratively work together to provide a more powerful feature representation capability than the individual adaptors. Figure~\ref{fig4_Adaptor} presents the overview of our collaborative adaptor, which consists of gated low-rank adaptation, prefix-tuning, and attentive fusion modules. Those adaptive modules are hierarchically incorporated into different levels of the Transformer architecture. 
In detail, the gated low-rank adaptor ($W_l$) is inserted after the bottom hidden state layer of transformer architecture. It maps the selective embedding features from previous layers. 
In the intermediate level of the Transformer architecture, the prefix-tuning ($W_p$) introduces additional trainable prefix tokens before the key $K$ and value $V$ to incorporate new task-specific information into the PLMs. 
At the top of Transformer architecture, we introduce the attentive fusion module ($W_f$) to further select task-related features from value $V$. Therefore, all adaptive modules collaborate to learn task-specific information and feature representations in a hierarchical manner. 

More specifically, the gated low-rank adaptation not only maintains the parameter efficiency of the low-rank adaptor (LoRA)~\cite{hu2022lora}, but also introduces a gate function $\Phi$ to selectively incorporate the learned features from LoRA. This gate function empowers the vanilla LoRA with a similar attentive capability to the Transformers module, by passing through the learned features selectively. Here, we default set the gated function $\Phi$ to $Sigmoid$ function. Mathematically, if we denote the output of fully connection layer with normalization as the hidden state $H$, the low-rank downscale metric $W_{down}$ and upscale metric $W_{up}$ map features in an efficient-computing manner. Our gated LoRA is defined as follows: 

\begin{equation}
   I^l = \Phi (W_{up}^T W_{down} H) + W_{I} H, I \in \{K,V,Q\}, 
    \label{Eq7:gLoRA}
\end{equation}

Gated LoRA selectively passes the learned features from LoRA with the gated function $\Phi$, which captures task-specific representation. The LoRA features are then collaborative with the Transformer's features to form the key $K$, value $V$, and query $Q$ through an additive operation correspondingly. Before feeding the mapped embedding $<K, V, Q>$ into multi-head computation, we introduce prefix tokens ($\mathtt{T}$) into query and key ($<K, V>$) to further enrich the feature learning capability of our detection model in the efficient-parameter learning. 

\begin{equation}
    Z_i = \mathtt{Attention}_i( Q, <\mathtt{T}_k; K>, <\mathtt{T}_v; V> ) ,
\end{equation}

$Z_i$ is the $i^{th}$ output of the multi-head attention computation. The prefix-tuning adaptors are introduced at the intermediate level of every Transformer, with significantly fewer trainable parameters compared to full parameter fine-tuning. To leverage the attentive mechanism, we introduce attentive fusion at the top of the Transformer, selectively activating the prefix-tuning features. As an attentive network, the key ($K'$) and value ($V'$) are derived from the outputs ($Z_i$) of the multi-head attention layer, and the query ($Q'$) is obtained from the previous layer's query ($Q$) through a residual connection. The attentive fusion is performed using dot production $\bigotimes$. All computations of attentive fusion are listed as follows:

\begin{equation}
\begin{split}
     Z_{af} &= \mathtt{Softmax}(Q' \bigotimes K'), \\
     Z      &= Z_{af}              \bigotimes V' .
\end{split}
\end{equation}

In this way, our collaborative adaptor performs efficient-parameter learning architecturally across bottom embedding layers, middle of Transformer, and top feature fusion. All the efficient modules collaborate with each other to learn a generic representation for the low-resource stance detection tasks. 

Besides the collaborative adaptor in efficient-parameter learning, we also introduce knowledge augmentation to facilitate efficient-parameter learning. 
In the knowledge infusion module, the crawled knowledge obtained from the Internet often exceeds the maximum input length of backbone PLMs. Our knowledge augmentation approach involves slicing the lengthy knowledge content into properly segmented parts to help the detection model capture the complete semantics of the infused knowledge. 

Unlike previous approaches in knowledge-infused stance detection that infuses knowledge ($\mathbb{K}$) following the paradigm 
\[ [ \; \mathbb{T},\, \mathbb{C}, <\mathtt{SEP}>,  \mathbb{K} , <\mathtt{CLS}>] \] 
or 
\[  [ \; \mathbb{T} <\mathtt{SEP}>, \mathbb{C}, <\mathtt{SEP}> , \mathbb{K},  <\mathtt{CLS}>] \] 
where $\mathbb{T}$, $\mathbb{C}$, and $\mathbb{K}$ denote the target, given context, and knowledge, respectively. $<\mathtt{SEP}>$ and $<\mathtt{CLS}>$ are separate token and ending token for the PLMs. 
we reformulate our knowledge infusion paradigm into 
\[ [  \mathbb{P}_t, \mathbb{C} , <\mathtt{SEP}> , \mathbb{K}_{sub} , <\mathtt{CLS}> ] \] 
Our input paradigm employs the prompt $\mathbb{P}_t$: 
\[ ``What's ~ the ~ stance ~ of ~ \mathbb{T} ~ in ~ following ~ context?'' \]
instead of the target $\mathbb{T}$ to fully leverage the capability of PLMs, which matches the pre-training input format of two sentences split by $<\mathtt{SEP}> $, as well as keeping the semantic integrity. 

In detail, we conduct the knowledge augmentation by slicing the long collaborative knowledge content into sub-knowledge segments, each of which fits the maximum length requirement. This manner helps the stance detection model capture the entire background knowledge instead of the cropped knowledge with missing information, as in the previous knowledge enhancement paradigms. The sub-knowledge segment $\mathbb{K}_{sub}$ is sampled from the collaborative knowledge $\mathbb{K}$ as follows:

\begin{equation}
\begin{split}
    \mathbb{K}_{sub}   & = [\mathbb{K}_{i*{l}/{2}},\; \mathbb{K}_{(i+1)*{l}/{2}}],  \\
    \mbox{\textit{s.t.}} \quad & i \in (0, 1, ..., \left \lfloor {\mathtt{Len}(\mathbb{K})}/{l}\right \rfloor ).
\end{split}
\end{equation}

The collaborative adaptor and knowledge augmentation work together to optimize efficient-parameter learning by reducing trainable parameters and addressing data limitations in low-resource stance detection tasks. The collaborative adaptor reduces the data consumption in training large-scale PLMs, while knowledge augmentation expands the training data to further improve training efficiency.

\subsection{Staged Optimization Algorithm}
To address the challenges of data discrepancy and domain gap between training data and pretrained models in the low-resource stance detection task, we propose a staged optimization algorithm that combines collaborative knowledge infusion and efficient parameter learning. However, the collaborative adaptor weights are initialized randomly, which may hinder its cooperation with the pretrained backbone PLMs during the initial training phase. Another issue is the unbalanced data distribution that is often overlooked in stance detection tasks. Our algorithm aims to mitigate these challenges in low-resource stance detection.

\begin{algorithm}
\caption{Staged Optimization Algorithm}
\label{alg:stagedAlgorithm}
\begin{algorithmic}[1]
\Require
\Statex Initialize the target's collaborative knowledge $\mathbb{K}$.
\Statex Initialize the weights of collaborative adaptor $W_{l,p,f}$.
\Ensure
\Statex Set the input $<\mathbb{C}, \mathbb{T}, \mathbb{K}>$ and stance label $Y$. 
\Statex Set the total steps $N$, stage-1 step $N_{s1}$, stage-2 steps $N_{s2}$.
\Statex Set the initial learning rate $\alpha_0$, Set the loss weights [$\theta_a, \theta_b$].
\Repeat
  \For{$it = 1$ to $N$}
    \State Fetch training data batch $<\mathbb{C}, \mathbb{T}, Y>$.
    \State Set knowledge augmentation $\mathbb{K}_{sub} \gets <\mathbb{T},\mathbb{K}>$.
    \State Set Prompt $\mathbb{P} \gets \mathbb{P}_t$.

    \State \textbf{Stage-1: label smooth}
    \If{$it \leq N_{s1}$}   
        \State Set smooth factor $\varepsilon$
    \Else
        \State $\varepsilon \gets 0$
    \EndIf

    \State \textbf{Stage-2: weighted loss}
    \If{$it \geq N_{s2}$}      
        \State  $\theta \gets \theta_a $
    \Else
        \State $\theta \gets \theta_b $
    \EndIf
    
    \State \textbf{Feed Forward computing:}
    \State \qquad $\tilde{Y} \gets  model(\mathbb{C},\mathbb{P}_t,\mathbb{K}_{sub}|W, W_{l,p,f}) $
    \State \textbf{Compute cost function:}
    \State \qquad $ CE(Y,\tilde{Y}) \gets -\sum_{c=1}^{c} \theta \cdot Y_c log(\tilde{Y}_c)$  
    \State $ J(Y,\tilde{Y}) \gets -(1-\varepsilon)log(1-CE(Y,\tilde{Y})-\varepsilon log(CE(Y,\tilde{Y})) $
    \State \textbf{Compute gradient:}
    \State \qquad $ \Delta W_{l,p,f} \gets \frac{\partial J(Y,\tilde{Y}) }{ \partial W_{l,p,f} } $         
    \State \textbf{Update gradient:}
    \State \qquad $ W_{l,p,f} \gets W_{l,p,f} + \alpha \Delta W_{l,p,f} $  
    \EndFor    
\Until{convergence}
\end{algorithmic}
\end{algorithm}

Algorithm~\ref{alg:stagedAlgorithm} presents the algorithm details for the low-resource stance detection task. 
Once the collaborative knowledge $\mathbb{K}$ and adaptor initialization are prepared, we set the first stage step for label smooth training, and set the second stage with the weighted loss for unbalanced stance categories. Before training begins, we prepare the prompt $\mathbb{P}_t$ and augmented knowledge $\mathbb{K}_{sub}$ with the given stance target $\mathbb{T}$ as the training input triplet $<\mathbb{C}, \mathbb{P}_t, \mathbb{K}_{sub}>$. Our two-stage learning algorithm incorporates weighted loss and label smoothing to improve the adaptive model's performance in unbalanced stance detection. Our algorithm aims to enhance the model's capability to handle ambiguous and diverse inputs by adjusting the loss function weights and introducing label smoothing during the training process. 
In the first stage, label smoothing is applied to soften the training targets, allowing the model to better handle uncertain data instances. Meanwhile, label smoothing helps the collaborative adapter convergence with newly initialed parameters. 
In the second stage, a weighted loss function is employed to assign different weights to different classes, thereby improving the model's ability to handle unbalanced datasets. Our algorithm provides a promising solution for tackling the challenges of data discrepancy and unbalanced data distribution in low-resource stance tasks.

\section{Experiments}
To evaluate the effectiveness of our proposed method, we conducted extensive experiments on three publicly available stance detection datasets and different low-resource settings. In this section, we provide a brief description of the datasets and the compared methods used in our stance detection task. Finally, we summarize the results using F1 as the default evaluation metric.

\subsection{Datasets}
\textbf{VAST}~\cite{allaway-mckeown-2020-zero} is a typical zero-shot/few-shot stance detection dataset that covers a wide range of over $6000$ targets across various themes, including politics, sports, education, immigration, and public health, etc. The \textit{VAST} dataset consists of $13447, 2062, 3006$ examples in its training, validation, and test sets, respectively. Notably, the majority of targets in \textit{VAST} are designed for zero-shot setting. It has an average of approximately $2.4$ examples per target. This characteristic makes \textit{VAST} particularly suitable for zero-shot/few-shot stance detection tasks. 

\textbf{P-Stance}~\cite{li-etal-2021-p} is stance detection dataset specific to the political domain. It contains in-target and cross-target settings with $21,574$ labeled tweets on three specific targets: ``Biden'', ``Sanders'' and ``Trump''. In the in-target setting, the target and classifier are the same in both the training and evaluation sets. Conversely, in the cross-target setting, the targets are entirely different, allowing for the evaluation of the generalization performance. 

\textbf{COVID-19-Stance}~\cite{glandt-etal-2021-stance} is stance detection dataset constructed from COVID-19-related tweets. It contains $6,133$ tweets with respect to four specific targets: ``Anthony S. Fauci, M.D. (Fauci)'', ``Keep School Closed (School)'', ``Stay at Home Order (Home)'' and ``Wearing a Face Mask (Mask)''. \textit{COVID-19-Stance} is also an unbalanced dataset in terms of class distribution.

\subsection{Compared Methods} 

\textit{TAN}~\cite{ijcai2017p557} is a classical attention-based method for the stance detection task. It contains a target-specific attention extractor and a long short-term memory network. 

\textit{BERT}~\cite{devlin-etal-2019-bert} is a well-known Transformer-based pretrained language model widely used for various downstream tasks. We employ BERT as our baseline for reference in the stance detection task. 

\textit{WS-BERT-Dual}~\cite{he-etal-2022-infusing} infuses target-related knowledge from extra Wikipedia to enhance background knowledge of PLMs in stance detection tasks. It utilizes two pretrained encoders to encode tweets and knowledge separately.

In addition to the shared baselines mentioned above, we also introduce other strong specific baselines for different stance datasets. For VAST task, we introduce graph convolution networks-based methods \textit{BERT-GCN}, \textit{CKE-NET}~\cite{lin-etal-2021-bertgcn}, and \textit{BSRGCN}~\cite{luo-etal-2022-exploiting}. Those methods joined large pretrained models with graphic convolution networks to leverage the learning capability of the models on heterogeneous data with structured graphic representation. For the zero-shot setting in \textit{VAST} task, we select BSRGCN and contrastive learning-based \textit{Joint-CL}~\cite{liang-etal-2022-jointcl} as the compared methods.

For \textit{PStance} task, we choose the bi-recurrent neural networks (\textit{BiCE})~\cite{augenstein-etal-2016-stance} and gated convolutional neural networks (\textit{GCAE})~\cite{xue-li-2018-aspect} and PGCNN~\cite{huang-carley-2018-parameterized} as the baselines. Specifically, the GCAE uses a TanH as the gate function to selectively output the sentiment futures according to the given aspect. Similarly, PGCNN also used the parameterized filters as the gate function to effectively capture the aspect-specific features.  Moreover, we also include \textit{BERT-Tweet}~\cite{li-etal-2021-p} as the compared method, which is pretrained on the target Tweet domain data.  

For \textit{COVID-19-Stance} task, we choose \textit{CT-BERT}~\cite{glandt-etal-2021-stance} as the baseline, which pretrained on the COVID-19-related tweet corpus to enhance task-specific domain knowledge. We also include \textit{CT-BERT-NS} and \textit{CT-BERT-DAN}~\cite{zhang-etal-2020-enhancing-cross}, which incorporate self-training and domain adaption into CT-BERT to further improve model representation capability and reduce domain gap in the stance detection task.

\subsection{Implementation Details}
We utilize RoBERTa as the backbone model for both \textit{VASE} and \textit{PStance} tasks. The batch size is set to $16$, and the learning rates range from $1e-5$ to $5e-5$ in our experiments. Since \textit{COVID-19-Stance} is a task related to the COVID-19 pandemic, we employ CT-BERT and BERT as the backbone models to leverage COVID-19-related knowledge from pretrained models. Due to our GPU memory limitations, the batch size is set to $8$. Regarding the hyperparameters of the collaborative adaptor, we set the rank to $r=8$ for the low-rank adaptor. The prefix-taken is set to $100$ with a dropout rate $0.2$. The reducing factor for the feature fusion module is $16$, and all gates are the \textit{ReLU} function. The models are implemented using Pytorch, and the maximum input length is default set to $512$ tokens. We trained the models for a maximum of $30$ epochs, and applied stopping with a patience of $10$ epochs. The optimizer used is AdamW with a weight decay of $1e-5$. All the experiments are conducted with the same random seed on four \textit{NVIDIA RTX A5000} GPU cards.

\subsection{Results}
To verify the effectiveness of our proposed method, we evaluate its performance on three public stance detection tasks: \textit{VAST}, \textit{PStance}, and \textit{COVID-19-Stance}. Firstly, we evaluate the proposed method on the \textit{VAST}, a low-resource dataset with a significantly larger number of targets than the other two datasets. Additionally, we evaluate the method's performance on the \textit{PStance} and \textit{COVID-19-Stance} datasets. Following previous work~\cite{he-etal-2022-infusing}, all the datasets are evaluated using the macro-average F1-score as the standard metric. The overall performance is calculated as the average across all stances.

\textit{VAST} dataset officially splits into two sub-tasks: zero-shot stance detection (\textit{600} targets) and few-shot stance detection (\textit{159} targets). Zero-shot setting does not include any targets in its training set, while the few-shot setting has very limited training samples (approximately $14.8$ examples per target) in its training set. In contrast, the \textit{PStance} and \textit{COVID-19-Stance} datasets have over hundreds of training examples per target. Table~\ref{tab:VAST} summarizes the evaluation results on the \textit{VAST} dataset. 

\begin{table}[!ht]
\caption{Experimental Results on VAST}
\label{tab:VAST}
\centering
\begin{tabular}{lccc}
\hline
\textbf{Method} & \textbf{Zero-shot} & \textbf{Few-shot} & \textbf{Overall} \\ \hline
TAN          & 66.6 & 66.3 & 66.5 \\
BERT         & 68.5 & 68.4 & 68.4 \\
BERT-GCN     & 68.6 & 69.7 & 69.2 \\
CKE-Net      & 70.2 & 70.1 & 70.1 \\
BSRGCN       & 72.6 & 70.2 & 71.3 \\
Joint-CL     & 72.3 & 71.6 & 72.3 \\
WS-BERT-Dual & 75.3 & 73.6 & 74.5 \\    \hline
\textbf{Ours} & \textbf{81.9} & \textbf{79.6} & \textbf{80.7} \\ \hline
\end{tabular}
\end{table}

From the numbers presented in Table~\ref{tab:VAST}, we can observe that the baseline method BERT achieves clear improvements around $2\%$ in the overall performance compared to the none pretrained baseline TAN, which indicates pretrained models have a stronger feature representation capability than none pretrained TAN.  Building upon BERT, BERT-GCN incorporates graphic convolution networks (GCN) with BERT further improving the overall F1 score to $69.2$. Similarly, the GCN-based methods CKE-Net and BSRGCN demonstrate progressive improvements by leveraging graph convolution networks, achieving an F1 score of $71.3$. Specifically, BSRGCN performs better in the zero-shot setting, benefiting from the unsupervised training on the domain-specific corpus. Joint-CL further enhances model performance through contrastive learning, achieving an overall F1-score of $72.3$. To enhance background knowledge, WS-BERT-Dual introduces target-related knowledge from Wikipedia, resulting in signification improvements compared to previous methods. 

However, all those solutions overlooked the fact that \textit{VAST} is a low-resource task, particularly for large pretrained models. Our method addresses this issue by incorporating efficient-parameter learning, and staged optimization for the low-resource task. Another neglect point in those solutions is that the infused target's knowledge should be the corrected knowledge. Our method addresses this issue by incorporating the collaborative knowledge infusion that introduces knowledge from multiple knowledge sources in a more accurate way. As a result, our method achieves new state-of-the-art (SOTA) performance on \textit{VAST}, achieving an overall F1-score of $80.7$. Interestingly, we find that the zero-shot settings achieve higher scores than the few-shot settings, especially in the pretrain-based methods. This difference can be attributed to the fact that the zero-shot and few-shot sets are two distinct subsets with completely different targets in the test set. Consequently, we can treat these two settings as two separate datasets.

\begin{table}[!ht]
\caption{Experimental Results on PStance}
\label{tab:PStance}
\centering
\begin{tabular}{lcccc}
\hline
\textbf{Method} & \textbf{Trump} & \textbf{Biden} & \textbf{Sanders} & \textbf{Avg.} \\ \hline
TAN         & 77.1      & 77.6      & 71.6      & 75.1      \\
BiCE        & 77.2      & 77.7      & 71.2      & 75.4       \\
PGCNN       & 76.9      & 76.6      & 72.1      & 75.2       \\
GCAE        & 79.0      & 78.0      & 71.8      & 76.3       \\
BERT        & 78.3      & 78.7      & 72.5      & 76.5       \\
BERT-Tweet  & 82.5      & 81.0      & 78.1      & 80.5       \\
WS-BERT-Dual & 85.8     & 83.5      & 79.0      & 82.8       \\ \hline
\textbf{Ours} & \textbf{86.2} & \textbf{84.1} & \textbf{80.5} & \textbf{83.6} \\ \hline 
\end{tabular}
\end{table}

Different from VAST with a large number of targets in a low-resource setting, \textit{PStance} contains only $3$ targets, and \textit{COVID-19-Stance} contains $4$ targets. 
Table~\ref{tab:PStance} presents the evaluation results of compared methods on \textit{PStance}. The classical recurrent neural network-based TAN and BiCE obtain comparable performance to the GCN-based PGCNN and GCAE, yielding an F1-score around $75~76$. Those results approach the performance of pretrained BERT. This similarity in performances suggests that rich training sources can benefit different types of models, and even achieve comparable performance to pretrain-based BERT. Different from BERT pretrained on the generic corpus, BERT-Tweet enhances domain-specific knowledge by being pretrained on the Twitter corpus, resulting in significant improvements of $4\%$ in the overall F1-score. Building upon BERT-Tweet, WS-BERT-Dual further infused the target background knowledge from Wikipedia, attaining an overall F1-score of $82.8$. In light of WS-BERT-Dual, our method further optimizes knowledge augmentation through parameter-efficient learning, achieving the best performance across all the targets than the compared methods in \textit{PStance} task.

\begin{table}[!ht]
\caption{Experimental Results on COVID-19-Stance}
\label{tab:Covid19}
\centering
\begin{tabular}{lccccc}
\hline
\textbf{Method} & \textbf{Fauci} & \textbf{Home} & \textbf{Mask} & \textbf{School} & \textbf{Avg.} \\ \hline
TAN             & 54.7 & 53.6 & 54.6 & 53.4 & 54.1 \\
ATRGU           & 61.2 & 52.1 & 59.9 & 52.7 & 56.5 \\
GCAE            & 64.0 & 64.5 & 63.3 & 49.0 & 60.2 \\
CT-BERT         & 81.8 & 80.0 & 80.3 & 75.3 & 79.8 \\
CT-BERT-NS      & 82.1 & 78.4 & 83.3 & 75.3 & 79.8 \\
CT-BERT-DAN     & 83.2 & 78.7 & 82.5 & 71.7 & 79.0 \\
WS-BERT-Dual    & 83.6 & 85.0 & 86.6 & 82.2 & 84.4 \\ \hline
\textbf{Ours} & \textbf{86.05} & \textbf{86.76} & \ \textbf{86.91} & \textbf{83.33} & \textbf{85.76} \\ \hline 
\end{tabular}
\end{table}

We also conducted evaluations of the proposed method on the domain-specific COVID stance detection and present the results in Table~\ref{tab:Covid19}. From the comparison of the results, we can clearly observe the performance gap between traditional gated-based methods (TAN, ATGRU, GCAE) and pretrain-based models (CT-BERT and its variants). The gated-based methods, which only conduct finetuning on its rich training set, obtain low average F1-scores below $60.2$, lacking any background knowledge specific to the target domain. 

In contrast, pretrained models trained on the COVID-related Twitters data exhibit good background knowledge and feature representation for \textit{COVID-19-Stance}, resulting in a high average F1-score above $79.0$. Meanwhile, self-training and domain adaptation techniques applied to CT-BERT led to performance improvements in stance detection for ``Fauci'' and ``Mask'', but no substantial improvements in the overall F1-score. With the help of dual pretrained model encoders, WS-BERT-Dual further elevates the overall performance to $84.4$. Similarly, our method achieves the best performance among the compared methods in \textit{COVID-19-Stance} by leveraging collaborative adaptor and staged optimization. Based on the extensive experimental comparisons, we can conclude that our proposed method performs well not only in low-resource \textit{VAST} stance detection task but also in rich-resource \textit{PStance} and \textit{COVID-19-Stance} tasks.

\section{Discussion}

\subsection{Ablation Study}
We performed an ablation study on the main modules of our method, namely collaborative knowledge infusion, efficient parameter learning, and staged optimization, using the \textit{VAST} and \textit{PStance} datasets. In addition to reporting the overall F1-score for zero-shot and few-shot settings in \textit{VAST}, we also provide the detailed results for three specific stances: pros, cons, and neutral. For \textit{PStance}, we report the average performance across the different targets.

\begin{table*}[!th]
\centering
\begin{threeparttable}
\caption{Ablation Study on VAST}
\label{Tab_Ablation_VAST}
\begin{tabular}{rcccccccccccc}
\hline
\textbf{Backbones} & \textbf{KI} & \textbf{EP} & \textbf{SO} & \textbf{Cons(ZS)} & \textbf{Pros(ZS)} & \textbf{Neu(ZS)} & \textbf{Zero-Shot} & \textbf{Cons(FS)} & \textbf{Pros(FS)} & \textbf{Neu(FS)} & \textbf{Few-Shot} & \textbf{Avg.} \\ \hline
RoBERTa-L &            &             &             & 65.7 & 59.0 & 95.0  & \multicolumn{1}{c|}{73.3} & 65.6 & 60.5 & 97.4 & \multicolumn{1}{c|}{74.5} &  73.9  \\
RoBERTa-L &\Checkmark  &             &             & 66.4 & 64.9 & 93.5  & \multicolumn{1}{c|}{74.9} & 66.4 & 63.2 & 96.3 & \multicolumn{1}{c|}{75.3} &  75.1  \\
RoBERTa-L &            & \Checkmark  &             & 71.4 & 72.3 & 90.7  & \multicolumn{1}{c|}{78.1} & 67.7 & 68.6 & 88.5 & \multicolumn{1}{c|}{74.9} &  76.5  \\
RoBERTa-L &            &             & \Checkmark  & 74.5 & 72.0 & 92.0  & \multicolumn{1}{c|}{79.5} & 69.2 & 67.8 & 87.0  & \multicolumn{1}{c|}{74.7} &  77.0  \\
RoBERTa-L &\Checkmark  & \Checkmark  &             & 70.7 & 70.4 & 95.5  & \multicolumn{1}{c|}{78.9} & 68.9 & 70.1 & 95.9 & \multicolumn{1}{c|}{78.3} &  78.6  \\
RoBERTa-L &            & \Checkmark  & \Checkmark  & 69.1 & 73.5 & 94.9  & \multicolumn{1}{c|}{79.2} & 68.5 & 72.8 & 96.7 & \multicolumn{1}{c|}{79.3} &  79.3  \\ \hline
Ours &\Checkmark  & \Checkmark  & \Checkmark  & 75.2 & 75.3 & 95.1  &\multicolumn{1}{c|}{\textbf{81.9}} & 71.5 & 72.6 & 94.7 &  \multicolumn{1}{c|}{\textbf{79.6}} &  \textbf{80.7}  \\ \hline
\end{tabular}
\begin{tablenotes}
    \small
    \item `Neu' is short for the neutral stance.
    \item `ZS' and `FS' are shorts of Zero-shot and Few-shot. `L' denotes the large size backbone.
    \item `KI', `EP', and `SO' denote knowledge infusion, collaborative adaptor, and staged optimization.
   \end{tablenotes}
\end{threeparttable}
\end{table*}

Table~\ref{Tab_Ablation_VAST} presents the results of the ablation study conducted on VAST, where RoBERTa serves as the backbone model. We study the impact of each module on the backbone performance. All individual modules that work with the backbone outperformed the finetuning of the vanilla backbone. The collaborative knowledge infusion (\textit{KI}) module, which includes knowledge verification and augmentation, facilitated the learning of target-specific background knowledge and achieved remarkable improvements of $75.1\%$. Likewise, the efficient parameter learning module (\textit{EP}) proved beneficial for the stance detection model on the low-resource \textit{VAST} data with the help of collaborative adaptors. 
When comparing the performance across different stances, we observed that the neutral stance exhibited significantly higher scores compared to the pros and cons stances. Our staged optimization (\textit{SO}) module tries to address this bias by incorporating label smooth and weighted loss, resulting in overall performance improvements. Furthermore, extensive ablation studies were conducted to assess different combinations of the modules. We can observe that two module combinations further improve model performance by $1\%-2\%$. Similarly, our method incorporating all three modules achieved the best overall F1-score of  $80.7\%$ on \textit{VAST}, highlighting the effectiveness of each module in addressing the challenges of the low-resource \textit{VAST} task.

\begin{table}[!ht]
\caption{Ablation Study on PStance}
\label{Tab:Ablation_PStance}
\centering
\begin{threeparttable}
\begin{tabular}{lccccccc}
\hline
\textbf{Setting}& \textbf{KI} & \textbf{EP} & \textbf{SO} & \textbf{Trump} & \textbf{Biden} & \textbf{Sanders} & \textbf{Avg.} \\ \hline
0&               &                &               & 85.5  & 82.7  & 76.9    & 81.7 \\
1&\Checkmark     &                &               & 84.6  & 82.7  & 76.9    & 81.4 \\
2&               &  \Checkmark    &               & 76.3  & 80.5  & 78.5    & 78.4 \\
3&               &                &  \Checkmark   & 84.4  & 83.6  & 78.4    & 82.1 \\
4&\Checkmark     & \Checkmark     &               & 84.8  & 82.7  & 73.7    & 80.4 \\
5&               &  \Checkmark    &  \Checkmark   & 86.1  & 83.3  & 80.5    & 83.3 \\
6&\Checkmark     &                & \Checkmark    & 85.0  & \textbf{85.5 } & \textbf{80.5}    & \textbf{83.7} \\ \hline
7&\Checkmark     & \Checkmark     & \Checkmark    & \textbf{86.2}  & 84.1  & \textbf{80.5}    &  83.6  \\ \hline
\end{tabular}
\begin{tablenotes}
\item `KI', `EP', and `SO' denote collaborative knowledge infusion, efficient parameter learning, and staged optimization.
\end{tablenotes}
\end{threeparttable}
\end{table}

We also conducted an ablation study on \textit{PStance}, which benefits from a relatively rich training source than the low-resource \textit{VAST} dataset. Table~\ref{Tab:Ablation_PStance} presents a summary of the ablation study on \textit{PStance}, focusing on the efficiency of three modules with the same backbone RoBERTa. Notably, the ablation study results differ from the results obtained for VAST. 
Interestingly, we observed a slight decrease in model performance with the collaborative knowledge infusion (KI) module than the vanilla backbone. This performance drop may be attributed to the specific dataset, as \textit{PStance} only contains four targets compared to diverse targets in \textit{VAST}. In other words, only four background long-sequential knowledge content $\mathbb{K}$ is infused into the model training pipeline, which may hinder feature learning on short-sequential raw tweets content $\mathbb{C}$. 
In the single-modular settings, we found backbone incorporating knowledge infusion (KI) or staged optimization (SO) exhibited superior performance compared to the efficient parameter learning (EP) module in the rich-source \textit{PStance} task. This suggests that full parameter finetuning is more effective than the adaptor-based solution in data-rich tasks. Similarly, in the two-modular settings, we observed that the setting with EP module (setting 4\&6) performed worse than the setting without EP modular (setting 5). We also observe that the backbone with two-modular settings yielded more improvements in the overall F1-score than the single-modular settings. In the three-modular setting, we find the performance of the `Trump' and `Sanders' targets could be further improved compared to the two-modular settings. However, the `Biden' category experienced a significant drop compared to its results in the setting-6. We attribute this to the negative impact of the `EP' module in the three-modular setting, which slightly decreased the overall performance compared to the two-modular setting 6. Thus, we can conclude that the adaptor-based solution does not always perform well in rich-resource tasks. 

\subsection{Cross-target Stance Detection on PStance}
We evaluate the model's generalization performance on the rich-source \textit{PStance}, we conducted cross-target stance detection, training model on one target, and evaluated on another target (\textit{e.g.}, training on Trump and testing on Biden). We employed BERT-Tweet~\cite{li-etal-2021-p} and WS-BERT-Dual~\cite{he-etal-2022-infusing} as the strong baselines. BERT-Tweet is pretrained on the Twitter corpus, benefiting from domain-specific knowledge. WS-BERT-Dual is a dual-path architecture using BERT and BERT-Tweet as feature encoders to incorporate target-specific Wikipedia knowledge. We follow the experimental settings of WS-BERT-Dual, testing on three targets: Trump, Biden, and Sanders. 
We employ the knowledge infusion and the staged optimization modules for cross-target stance detection tasks, as our ablation study demonstrated the two-modular setting performs best in rich-resource \textit{PStance}. All the experimental results are summarized in the following table:

\begin{table}[!ht]
\caption{Cross-target Stance Detection on PStance}
\label{Tab:Cross_PStance}
\centering
\begin{threeparttable} 
\begin{tabular}{rccc}
\hline
\textbf{Cross-Targets}      & \textbf{BERT-Tweet} & \textbf{WS-BERT-Dual} & \textbf{Ours }\\ \hline
$\mathtt{Trump} \rightarrow \mathtt{Biden}$     & 58.9 &  \textbf{68.3}        &  68.2     \\
$\mathtt{Trump} \rightarrow \mathtt{Sanders}$& 56.5 &  64.4   &  \textbf{67.8}    \\
$\mathtt{Biden} \rightarrow  \mathtt{Trump}$    & 63.6 &  67.7  &  \textbf{72.1 }   \\
$\mathtt{Biden } \rightarrow \mathtt{Sanders}$  & 67.0 &  69.0  &  \textbf{74.8 }   \\
$\mathtt{Sanders} \rightarrow \mathtt{Trump} $  & 58.7 &  \textbf{63.6}       &  63.4      \\
$\mathtt{Sanders} \rightarrow \mathtt{Biden} $  & 73.0 &  76.8   &  \textbf{78.9 }   \\ \hline
\textbf{\textit{Avg.}}      & 63.0          &  68.3 &  \textbf{70.9 }   \\ \hline
\end{tabular}
\begin{tablenotes}
\item[*] All the reported results are F1 score.
\item[*] In cross-target setting, the stance detection model is trained on left-target data and tested on the right-target data. 
\end{tablenotes}
\end{threeparttable}
\end{table}

From the results of Table~\ref{Tab:Cross_PStance}, we observed that the WS-BERT-D achieved significant improvements (average $5\%$ F1-score) in all six cross-target pairs compared to the baseline BERT-Tweet, benefiting from the additional BERT branch for encoding extra Wikipedia knowledge. In contrast, our method only uses the RoBERTa as the backbone for cross-target stance detection. We further improved four of six cross-target pairs by introducing the staged optimization, resulting in a slight performance drop in ``Trump$\rightarrow$ Biden'' and ``Sanders $\rightarrow$ Trump'' than the WB-BERT-Dual. Additionally, we noticed that the two pairs' target results are not symmetric to each other. Overall, our proposed solution achieved a new state-of-the-art performance in cross-target stance detection on the \textit{PStance} dataset.

\subsection{Low-resource Stance Detection}
In this section, we further evaluated our method's performance in the low-resource settings of stance detection on the subsets of PStance and COVID-19-Stance datasets. Specifically, we randomly sampled $5\%, 10\%, 15\%, 20\%$ data from their training sets as our low-resource data settings, and kept its test sets for evaluation. 

\begin{table}[htbp]
\caption{Low-resource Stance Detection on PStance}
\label{Tab:LR_PStance}
\begin{tabularx}{\linewidth}{XXXXXXXX}
\toprule
\multirow{2}{*}{\textbf{Settings}} & \multicolumn{3}{c}{\textbf{Baseline}} & \multicolumn{3}{c}{\textbf{Ours}} & \multirow{2}{*}{\textbf{Avg.($\%$) }} \\
\cmidrule(lr){2-4} \cmidrule(lr){5-7}
 & Trump & Biden & Sanders & Trump & Biden & Sanders & \\
\midrule
\textit{5$\%$} & 69.3 & 70.5 & 63.5 & 71.2 & 73.9 & 68.2 & 3.4 $\uparrow $ \\
\textit{10$\%$} & 70.5 & 72.4 & 71.0 & 73.8 & 75.0 & 72.5 & 2.5 $\uparrow$ \\
\textit{15$\%$} & 72.9 & 76.1 & 72.3 & 74.2 & 78.1 & 73.9 & 1.7 $\uparrow$ \\
\textit{20$\%$} & 74.0 & 78.4 & 74.5 & 76.9 & 79.4 & 75.1 & 1.7 $\uparrow$ \\
\bottomrule
\end{tabularx}
\begin{tablenotes}
\item[*] `Settings' means different low-resource settings with different sample percentages from PStance.
\item[*] `Avg.' denotes the average performance improved between our method and the baseline.
\end{tablenotes}
\end{table}

\begin{table*}[htbp]
\caption{Low-resource  Stance Detection on COVID-19-Stance}
\label{Tab:LR_COVID}
\centering
\begin{tabular}{cccccccccc}
\toprule
\multirow{2}{*}{\textbf{Settings}} & \multicolumn{4}{c}{\textbf{Baseline}} & \multicolumn{4}{c}{\textbf{Ours}} & \multirow{2}{*}{\textbf{Avg. ($\%$)}} \\ \cmidrule(lr){2-5} \cmidrule(lr){6-9}
              & \textbf{Fauci} & \textbf{Home} & \textbf{Mask} & \textbf{School} & \textbf{Fauci} & \textbf{Home} & \textbf{Mask} & \textbf{School} &      \\ \midrule
\textit{5$\%$}  & 34.6  & 44.0 & 37.0 & 22.7   & 40.8  & 44.0 & 37.0 & 22.7   & 6.6  $\uparrow$ \\
\textit{10$\%$} & 57.8  & 59.9 & 42.3 & 26.3   & 72.9  & 59.9 & 42.3 & 26.3   & 12.7 $\uparrow$ \\
\textit{15$\%$} & 69.6  & 62.1 & 49.0 & 40.3   & 77.2  & 62.1 & 49.0 & 40.3   & 6.4  $\uparrow$ \\
\textit{20$\%$} & 70.8  & 74.2 & 54.0 & 45.4   & 80.6  & 74.2 & 54.0 & 45.4   & 12.8 $\uparrow$ \\ \bottomrule
\end{tabular}
\end{table*}

Following the setting of the ablation study, we set the BERT as the baseline and backbone of our method in all the low-resource stance detection evaluations. Table~\ref{Tab:LR_PStance} and Table~\ref{Tab:LR_COVID} summarized the comparison results on different low-resource settings of PStance and COVID-19-Stance.  

From the reported results of Table~\ref{tab:PStance} on different low-resource settings, we can observe that our method surpasses the baseline compared to the baseline in all low-resource data settings, which shows our method presents the effectiveness of our proposed method. Overview the whole performance across different settings, we can obviously find the progressive increasing trends with the training data adding in both the baseline and our method. Last but not least, our method benefits more with less training data, and the average improvements are reduced with more data introduced from $5\%$ to $20\%$. 

Table~\ref{Tab:LR_COVID} summarized the low-resource setting evaluation results on the COVID-19-Stance dataset. Similar to the low-resource setting results of PStance, our method surpassed the baseline with large margins in all low-resource settings. The average performance can achieve more than $6\%$. Different from the performance trends in PStance, the performance improvement trends do not keep consistently changing with increasing of training data. We think the main reason causing this is the domain gap and diverse data distributions in different stance topics of COVID-19-Stance.


\subsection{Efficient Parameter Learning}
In this section, we compare the performance of our efficient parameter learning paradigm with the entire model finetuning paradigm in the low-resource stance detection task. We selected the classic BERT and RoBERTa for entire model finetuning, using both basic (B) and large (L) model sizes. Our efficient parameter learning approach utilizes the large-size RoBERTa as the backbone. All the models are evaluated in the zero-shot and few-shot settings as defined by \textit{VAST} dataset. Note, the few-shot and zero-shot in traditionally computer vision tasks are evaluated with same test data. However, in our VAST NLP dataset, the few-shot and zero-shot settings are evaluated with different test data.

\begin{table}[!ht]
\caption{Zero-shot Performance on VAST}
\label{Tab:LR_ZeroShort_VAST}
\centering
\begin{threeparttable}
\begin{tabular}{rccccc}
\hline
\textbf{Methods} & \textbf{Param$^*$}    & \textbf{Pros} & \textbf{Cons} & \textbf{Neutral} & \textbf{Avg.} \\ \hline
BERT-B     &\textit{110M}   &  64.0     &   63.2       &   94.2      &   73.8   \\
BERT-L     &\textit{340M}   &  60.0     &   66.8       &   94.1      &   72.4   \\
RoBERTa-B  &\textit{110M}   &  67.4     &   72.3       &   93.7      &   77.8  \\
RoBERTa-L  &\textit{340M}   &  65.7     &   59.0       &   95.0      &   73.3   \\ \hline
\textbf{Ours}       & \textit{3K}    &  75.2     &   75.3       &   95.1      &   81.9   \\ \hline
\end{tabular}
\begin{tablenotes}
\item Param$^*$ denotes the total trainable parameters. 
\item `B' and `L' denote the basic and large model sizes, respectively.
\end{tablenotes}
\end{threeparttable}
\end{table}

Table~\ref{Tab:LR_ZeroShort_VAST} presents the performance of different models on the zero-shot stance detection task with respect to three stances: pros, cons, and neutral. In the entire model finetuning setting, we observe that the basic-sized models outperformed the large-sized models in terms of average F1 score. Specifically, the BERT models experienced a $1\%$ drop in average F1-score from basic-size to large-size, while the RoBERTa models even encountered a decrease of more than $4\%$. 
The large model's zero-shot performance decay may attribute to combined training few-shot on the low-resource \textit{VAST} dataset. The large-size models exhibit reduced generalization capability with limited training data samples, resulting in performance decay in zero-shot stance detection tasks. Consistent with the findings of the ablation study, the neutral stance achieved significantly higher scores (above $90\%$) than the pros and cons stances in the zero-shot setting. In contrast, our efficient parameter learning method, which maintains the pretrained model's generalization capability by freezing its parameter, achieved the best performance in the zero-shot stance detection task.

\begin{table}[!ht]
\caption{Few-shot Performance on VAST}
\label{Tab:LR_FewShort_VAST}
\centering
\begin{threeparttable}
\begin{tabular}{rccccc}
\hline
\textbf{Methods} & \textbf{Param$^*$}    & \textbf{Pros} & \textbf{Cons} & \textbf{Neutral} & \textbf{Avg.} \\ \hline
BERT-B      &\textit{110M} &  64.2     &   65.1       &   95.1      &   74.8   \\
BERT-L      &\textit{340M} &  64.2     &   61.3       &   91.8      &   73.8   \\
RoBERTa-B   &\textit{110M} &  64.6     &   70.8       &   95.1      &   76.8   \\
RoBERTa-L   &\textit{340M} &  65.6     &   60.5       &   97.4      &   74.5   \\ \hline
\textbf{Ours}        &\textit{3K}   &  71.5     &   72.6       &   94.7      &   79.6   \\ \hline
\end{tabular}
\begin{tablenotes}
\item Param$^*$ denotes the total trainable parameters. 
\item `B' and `L' denote the basic and large model sizes, respectively.
\end{tablenotes}
\end{threeparttable}
\end{table}

Table~\ref{Tab:LR_FewShort_VAST} presents the results of different models in the few-shot VAST stance detection task. Similar to the zero-shot setting, we observe a decline in overall performance as the model size increases. Additionally, the neutral stance detection performance exhibited significant superiority over the pros and cons stances. For the entire model finetuning paradigm, there have been no notable variations between basic and large-size models in the pros and neutral stances. However, in the cons stance, the large-size models experienced a sharp drop. This can be attributed to the unbalanced data distribution, which leads to model performance decay when trained on limited samples. Therefore, the entire model finetuning using a large pretrained language model is not an optional solution for zero/few-shot stance detection tasks. The reason for zero-shot setting (Table \ref{Tab:LR_ZeroShort_VAST}) achieved better performance than the few-shot setting (Table \ref{Tab:LR_FewShort_VAST}) is mainly depends on the special test data setting in the VAST dataset that the few-shot and zero-shot settings have total different test data samples, which is different from traditional few-shot and zero-shot setting with the same test data. In other words, the settings in Table \ref{Tab:LR_ZeroShort_VAST} and \ref{Tab:LR_FewShort_VAST} can be seen as two datasets.  This divergence from traditional evaluation methods is a key factor in understanding the experimental results. Moreover, this pattern of zero-shot settings outperforming few-shot settings is not unique to our study but is also observed in comparable works, such as TAN [35], BERT [36], CKE-NET [9], BSRGCN [8], and WS-BERT-Dual [4], which report similar anomalies in Table 1. Our proposed method offers a solution for the low-resource stance detection task, which has only $3K$ trainable, significantly fewer than the millions of trainable parameters required for full model finetuning. Despite the parameter reduction, our method still can achieve superior performance with a considerable margin.

\subsection{Collaborative Adaptor Analysis}
Collaborative adaptor is an essential part of efficient parameter learning in the low-resource detection task, which consists of three modules: gated LoRA, prefix-tuning and attentive fusion. To assess the importance of each module, we conducted evaluations by removing individual modules from our collaborative adaptor. 

\begin{table}[!ht]
\caption{Collaborative Adaptor Analysis on VAST}
\label{Tab:Ablation_Adaptors}
\centering
\begin{threeparttable}
\begin{tabular}{ccccc}
\hline
\textbf{Settings}   & \textbf{Pros} & \textbf{Cons} & \textbf{Neutral} & \textbf{F1} \\ \hline
\textit{w/o} Gated LoRA        & 71.9           & 70.3             & 95.8             & 79.3        \\
\textit{w/o} Prefix-tuning     & 70.6           & 73.3             & 94.3             & 79.4       \\
\textit{w/o} Attentive Fusion  & 66.1           & 62.1             & 92.5             & 73.6   \\ \hline
\textbf{All}                   & 73.4           & 73.9             & 94.9             & 80.7        \\ \hline
\end{tabular}
\begin{tablenotes}
\item `\textit{w/o}' denotes without . 
\end{tablenotes}
\end{threeparttable}
\end{table}

Table~\ref{Tab:Ablation_Adaptors} presents the performance of different settings of the collaborative adaptor. For instance, the setting '\textit{w/o} gated LoRA' indicates the removal of the `gated LoRA' module. All settings were evaluated on the low-resource \textit{VAST} using the same hyperparameters. We observed the 'gated LoRA' and 'prefix-tuning' modules exhibited similar drops of approximately $1\%$ with slight variations across different stances. Surprisingly, in the `\textit{w/o} attentive fusion' setting, all stance scores experience a sharp decline of approximately $7\%$ on average. Through the performance comparisons, we discovered that the attentive fusion module had a more significant impact on down-steam stance detection tasks than the gated LoRA and Prefix-tuning modules in the efficient-parameter learning paradigm. One possible explanation for this observation is that attentive fusion is more closely connected to the stance prediction classifier than the other two models, which serve as feature extractors with less impact on the final stance prediction.


\subsection{Comparison with ChatGPT}
ChatGPT~\footnote{https://openai.com/blog/chatgpt} attracts lots of attention in the natural language processing community due to its impressive performance on conversational tasks, leading to its utilization in various downstream NLP tasks. In this section, we aim to evaluate the performance of ChatGPT on \textit{VAST}, which is a varied stance topics dataset with over a thousand targets. To adapt ChatGPT for the stance detection task, we constructed the prompt as follows: 

``\textit{Please choose one stance from $\mathtt{cons}, \mathtt{pros}, \mathtt{neutral}$ for $<\mathtt{TARGET}>(\mathbb{T})$ on following content: $<\mathtt{TWEET}>(\mathbb{C})$}?'' 

We sequentially selected and evaluated $100$ samples, comprising $33~\mathtt{cons}$, $33~\mathtt{pros}$, and $34~\mathtt{neutral}$ instances. Regarding ChatGPT is an evolving system\footnote{The ChatGPT results were evaluated on 04 June 2023.}, all the evaluation results are reported as follows, 

\begin{table}[!ht]
\caption{VAST Samples Evaluation on ChatGPT}
\label{Tab:ChatGPT}
\centering
\begin{threeparttable}
\begin{tabular}{rcccc}
\hline
\textbf{ GT/Pred} & \textbf{Cons} & \textbf{Pros}& \textbf{Neutral} & \textbf{\textit{Recall}} \\ \hline
\textbf{Cons} (\textit{33})  & 18           & 4            & 9            &  58.1     \\
\textbf{Pros}  (\textit{33})  & 3            & 18           & 14           &  51.4       \\
\textbf{Neutral}   (\textit{34})  & 12           & 11           & 11           &  33.3      \\ \hline
\textbf{\textit{Precision} }               & 54.5         & 54.5         & 32.4         &  /      \\ \hline
\end{tabular}
\begin{tablenotes}
\item The row is prediction (Pred) and the column is ground truth (GT).
\end{tablenotes}
\end{threeparttable}
\end{table}

The evaluation results presented in Table~\ref{Tab:ChatGPT} indicate that ChatGPT's performance on \textit{VAST} stance detection dataset is not as impressive as anticipated, which is similar to the findings of the work directly using the chain-of-thought~\cite{zhang2023investigating} in ChatGPT for stance detection on \textit{VAST} dataset with only $62.3$ F1 performance. From the result analysis, we observe that ChatGPT often predicts the pros or cons stances to the neutral stance, resulting in the neutral stance in a low recall score. This tendency might stem from ChatGPT's inclination to produce mild and friendly responses~\cite{wei2023simple}, leading to a bias toward predicting neutral stances. Furthermore,  we observed that ChatGPT trends to output a neutral stance for sensitive topics, such as voting, humanity, and elections. 


\section{Conclusion}
In this paper, we propose a method for low-resource stance detection that collaborative infuses verified target knowledge with efficient parameter learning. Firstly, we enhanced the infusion of target-related knowledge by extending it beyond structured Wikipedia to encompass a broader range of unstructured information from the entire Internet. To ensure the selection of relevant semantic background knowledge, a knowledge verifier is employed. Secondly, we introduce efficient-parameter learning through collaborative adaptors, which involve a minimal number of trainable parameters by freezing the weights of large PLM-based models. This manner not only facilitates efficient model training in low-resource stance detection tasks but also retains the rich prior knowledge encoded in pretrained models. Thirdly, a staged optimization algorithm is proposed to mitigate the impact of unbalanced data. Additionally, knowledge augmentation and prompting techniques are integrated into our efficient parameter learning framework for low-resource stance detection. Experimental results demonstrate the effectiveness of our method on three public datasets with state-of-the-art performance. In future work, we plan to further explore efficient-parameter learning in the context of multi-modal stance detection tasks.

\ifCLASSOPTIONcaptionsoff
  \newpage
\fi



\bibliographystyle{IEEEtran}
\bibliography{mybib}

%








\end{document}